\PassOptionsToPackage{unicode=true}{hyperref}
\documentclass{article}

\usepackage{cite}
\usepackage{hyperref}
\usepackage{tabularx}
\usepackage{graphicx,grffile}
\usepackage{amsmath}
\usepackage{amssymb}
\usepackage{bussproofs}
\usepackage[utf8]{inputenc}

\newtheorem{observation}{Observation}

\newtheorem{definition}{Definition}
\newtheorem{example}{Example}

\newenvironment{bprooftree}{\leavevmode\hbox\bgroup}{\DisplayProof\egroup}

\DeclareRobustCommand{\href}[2]{#2\footnote{\url{#1}}}

\begin{document}

\title{On Weakening Strategies for PB Solvers}

\author{
	Daniel Le Berre$^1$
	\and Pierre Marquis$^{1,2}$
	\and Romain Wallon$^{1}$
}

\date{
	$^1$CRIL, Univ Artois \& CNRS\\
	$^2$Institut Universitaire de France\\[2ex]
	\today
}

\maketitle
\begin{abstract}
Current pseudo-Boolean solvers implement different variants of the
cutting planes proof system to infer new constraints during conflict
analysis. One of these variants is \emph{generalized resolution}, which
allows to infer strong constraints, but suffers from the growth of
coefficients it generates while combining pseudo-Boolean constraints.
Another variant consists in using \emph{weakening} and \emph{division},
which is more efficient in practice but may infer weaker constraints. In
both cases, weakening is mandatory to derive conflicting constraints.
However, its impact on the performance of pseudo-Boolean solvers has not
been assessed so far. In this paper, new application strategies for this
rule are studied, aiming to infer strong constraints with small
coefficients. We implemented them in \emph{Sat4j} and observed that each
of them improves the runtime of the solver. While none of them performs
better than the others on all benchmarks, applying weakening on the
\emph{conflict} side has surprising good performance, whereas applying
\emph{partial} weakening and division on both the conflict and the
reason sides provides the best results overall.

\end{abstract}

\hypertarget{introduction}{%
\section{Introduction}\label{introduction}}

The last decades have seen many improvements in SAT solving that are at
the root of the success of modern SAT solvers
\cite{grasp, chaff, minisat}. Despite their practical efficiency on
many real-world instances, these solvers suffer from the weakness of the
resolution proof system they use in their conflict analyses.
Specifically, when proving the unsatisfiability of an input formula
requires an exponential number of resolution steps -- as for
pigeonhole-principle formulae \cite{haken85} -- a SAT solver cannot
find a refutation proof efficiently. This motivated the development of
pseudo-Boolean (PB) solvers \cite{handbook}, which take as input
conjunctions of \emph{PB constraints} (linear inequations over Boolean
variables) and apply \emph{cutting planes based inference} to derive
inconsistency \cite{gomory58, hooker88, nordstrom15}. This inference
system is stronger than the resolution proof system, as it
\emph{p-simulates} the latter: any resolution proof can be translated
into a cutting planes proof of polynomial size \cite{cook87}. Using
such a proof system may, in theory, make solvers more efficient: for
instance, a pigeonhole principle formula may be refuted with a linear
number of cutting planes steps.

However, in practice, PB solvers fail to keep the promises of the
theory. In particular, they only implement \emph{subsets} of the cutting
planes proof system, which degenerate to resolution when given a CNF
formula as input: they do not exploit the full power of the cutting
planes proof system \cite{jakobproofsolvers}. One of these subsets is
\emph{generalized resolution} \cite{hooker88}, which is implemented in
many PB solvers \cite{pbchaff, galena, pueblo, sat4j}. It consists in
using the \emph{cancellation} rule to combine constraints so as to
resolve away literals during conflict analysis, as SAT solvers do with
the resolution rule. Another approach has been introduced by
\emph{RoundingSat} \cite{rs}, which relies on the \emph{weakening} and
\emph{division} rules to infer constraints having smaller coefficients
to be more efficient in practice. These proof systems are described in
Section \ref{pseudo-boolean-solving}.

This paper follows the direction initiated by \emph{RoundingSat} and
investigates to what extent applying the \emph{weakening} rule may have
an impact on the performance of PB solvers. First, we show that applying
a \emph{partial} weakening instead of an aggressive weakening as
proposed in \cite{rs} allows to infer stronger constraints while
preserving the nice properties of \emph{RoundingSat}. Second, we show
that weakening operations can be extended to certain literals that are
falsified by the current partial assignment to derive shorter
constraints. Finally, we introduce a tradeoff strategy, trying to get
the best of both worlds. These new approaches are described in Section
\ref{weakening-strategies}, and empirically evaluated in Section
\ref{experimental-results}.

\hypertarget{pseudo-boolean-solving}{%
\section{Pseudo-Boolean Solving}\label{pseudo-boolean-solving}}

We consider a propositional setting defined on a finite set of
classically interpreted propositional variables \(V\). A \emph{literal}
\(l\) is a variable \(v \in V\) or its negation~\(\bar{v}\). Boolean
values are represented by the integers \(1\) (true) and \(0\) (false),
so that \(\bar{v} = 1 - v\). A \emph{PB constraint} is an integral
linear equation or inequation over Boolean variables. Such constraints
are supposed, w.l.o.g., to be in the normalized form
\(\sum_{i=1}^{n} \alpha_i l_i \geq \delta\), where \(\alpha_i\) (the
\emph{coefficients} or \emph{weights}) and \(\delta\) (the
\emph{degree}) are positive integers and \(l_i\) are literals. A
\emph{cardinality constraint} is a PB constraint with its weights equal
to \(1\) and a \emph{clause} is a cardinality constraint of degree
\(1\).

Several approaches have been designed for solving PB problems. One of
them consists in encoding the input into a CNF formula and let a SAT
solver decide its satisfiability \cite{minisatp, openwbo, naps}, while
another one relies on lazily translating PB constraints into clauses
during conflict analysis~\cite{satire}. However, such solvers are based
on the \emph{resolution} proof system, which is somewhat \emph{weak}:
instances that are hard for resolution are out of reach of SAT solvers.
In the following, we consider instead solvers based on the \emph{cutting
planes} proof system, the PB counterpart of the resolution proof system.
Such solvers handle PB constraints natively, and are based on one of the
two main subsets of cutting planes rules described below.

\hypertarget{generalized-resolution-based-solvers}{%
\subsection{Generalized Resolution Based
Solvers}\label{generalized-resolution-based-solvers}}

Following the CDCL algorithm of SAT solvers, PB solvers based on
\emph{generalized resolution} \cite{hooker88} make \emph{decisions} on
variables, which force other literals to be satisfied. These
\emph{propagated} literals are detected using the \emph{slack} of each
constraint.

\begin{definition}[slack]
Given a partial assignment $\rho$, the \emph{slack} of a constraint
$\sum_{i=1}^{n} \alpha_i l_i \geq \delta$ is the value
$\sum_{i=1, \rho(l_i) \neq 0}^{n} \alpha_i - \delta$.
\end{definition}

\begin{observation}
Let $s$ be the slack of the constraint $\sum_{i=1}^{n} \alpha_i l_i \geq \delta$
under some partial assignment.
If $s < 0$, the constraint is currently falsified.
Otherwise, the constraint requires all unassigned literals having a weight
$\alpha > s$ to be satisfied.
\end{observation}

\begin{example}
\label{ex:slack}
Let $\rho$ be the partial assignement such that $\rho(a) = 1$,
$\rho(c) = \rho(d) = \rho(e) = 0$ (all other variables are unassigned).
Under $\rho$, the constraint $6\bar{b} + 6c + 4e + f + g + h \geq 7$ has slack
$2$.
As $\bar{b}$ is unassigned and has weight $6 > 2$, this literal is propagated
(the constraint is the \emph{reason} for $\bar{b}$).
This propagation falsifies the constraint $5a + 4b + c + d \geq 6$, which now
has slack $-1$ (this is a \emph{conflict}).
\end{example}

When a conflict occurs, the solver analyzes this conflict to derive an
\emph{assertive} constraint, i.e., a constraint propagating some of its
literals. To do so, it applies successively the \emph{cancellation} rule
between the conflict and the reason for the propagation of one of its
literals (``LCM'' denotes the least common multiple):

\begin{prooftree}
\AxiomC{$\alpha l + \sum_{i=1}^{n} \alpha_i l_i \geq \delta$}
\AxiomC{$\beta \bar{l} + \sum_{i=1}^{n} \beta_i l_i \geq \delta'$}
\AxiomC{$\mu\alpha = \nu\beta = \text{LCM}(\alpha, \beta)$}
\RightLabel{(canc.)}
\TrinaryInfC{$\sum_{i=1}^{n} (\mu\alpha_i + \nu\beta_i) l_i \geq \mu\delta + \nu\delta' - \mu\alpha$}
\end{prooftree}

To make sure that an assertive constraint will be eventually derived,
the constraint produced by this operation has to be conflictual, which
is not guaranteed by the cancellation rule. To preserve the conflict,
one can take advantage of the fact that the slack is \emph{subadditive}:
the slack of the constraint obtained by applying the cancellation
between two constraints is at most equal to the sum of the slacks of
these constraints. Whenever the sum of both slacks is not negative, the
constraint may not be conflictual, and the \emph{weakening} and
\emph{saturation} rules are applied until the slack of the reason
becomes low enough to ensure the conflict to be preserved (only literals
that are not falsified may be weakened away).

\[\begin{bprooftree}
\AxiomC{$\alpha l + \sum_{i=1}^{n} \alpha_i l_i \geq \delta$}
\RightLabel{(weakening)}
\UnaryInfC{$\sum_{i=1}^{n} \alpha_i l_i \geq \delta - \alpha$}
\end{bprooftree}\qquad
\begin{bprooftree}
\AxiomC{$\sum_{i=1}^{n} \alpha_i l_i \geq \delta$}
\RightLabel{(saturation)}
\UnaryInfC{$\sum_{i=1}^{n} \min(\delta, \alpha_i) l_i \geq \delta$}
\end{bprooftree}\]

\begin{example}[Example~\ref{ex:slack} cont'd]
\label{ex:genres}
As $5a + 4b + c + d \geq 6$ is conflicting and $\bar{b}$ was propagated by
$6\bar{b} + 6c + 4e + f + g + h \geq 7$, the cancellation rule must be applied
between these two constraints to eliminate $b$.
To do so, the \emph{conflict side} (i.e., the first constraint) has to be
multiplied by $3$ and the \emph{reason side} (i.e., the second constraint) by
$2$, giving slack $-3$ and $4$, respectively.
As the sum of these values is equal to $1$, the resulting constraint is not
guaranteed to be conflicting.
Thus, the reason is weakened on $g$ and $h$ and saturated to get
$5\bar{b} + 5c + 4e + f \geq 5$, which has slack $1$.
To cancel $b$ out, this constraint is multiplied by $4$ and the conflict by
$5$, giving $25a + 25c + 16e + 5d + 4f \geq 30$, which has slack $-1$.
\end{example}

This approach has several drawbacks. Observe in Example \ref{ex:genres}
the growth of the coefficients in just one derivation step. In practice,
there are many such steps during conflict analysis, and the learned
constraints will be reused later on, so that coefficients will continue
to grow, requiring the use of arbitrary precision~arithmetic. Moreover,
after each weakening operation, the LCM of the coefficients must be
recomputed to estimate the slack, and other literals to be weakened must
be found. The cost of these operations motivated the development of
alternative proof systems, such as those weakening the derived
constraints to infer only cardinality constraints \cite{galena}, or
those based on the \emph{division} rule.

\hypertarget{division-based-solvers}{%
\subsection{Division Based Solvers}\label{division-based-solvers}}

To limit the growth of the coefficients during conflict analysis,
\emph{RoundingSat} \cite{rs} introduced an aggressive use of the
weakening and \emph{division} rules.

\begin{prooftree}
\AxiomC{$\sum_{i=1}^{n} \alpha_i l_i \geq \delta$}
\AxiomC{$r > 0$}
\RightLabel{(division)}
\BinaryInfC{$\sum_{i=1}^{n} \lceil \frac{\alpha_i}{r} \rceil l_i \geq \lceil \frac{\delta}{r} \rceil $}
\end{prooftree}

When a conflict occurs, both the conflict and the reason are weakened so
as to remove all literals not falsified by the current assignment and
having a coefficient not divisible by the weight of the literal used as
pivot for the cancellation, before being divided by this weight. This
ensures that the pivot has a weight equal to~\(1\), which guarantees
that the result of the cancellation will be conflictual~\cite{dixon04}.

\begin{example}[Example~\ref{ex:genres} cont'd]
\label{ex:div}
The weakening operation is applied on both the conflict $5a + 4b + c + d \geq 6$
and the reason $6\bar{b} + 6c + 4e + f + g + h \geq 7$, yielding
$4b + c + d \geq 1$ and $6\bar{b} + 6c + 4e \geq 4$, respectively.
Both constraints are then divided by the coefficient of the pivot ($4$ and $6$,
respectively), giving $b + c + d \geq 1$ and $\bar{b} + c + e \geq 1$.
Applying the cancellation rule on these two constraints gives
$2c + d + e \geq 1$, which is equivalent to the clause $c + d + e \geq 1$.
\end{example}

The \emph{RoundingSat} approach succeeds in keeping coefficients small,
and its aggressive weakening allows to find the literals to remove
efficiently. However, some constraints inferred by this solver may be
weaker than those inferred with generalized resolution (compare the
constraints derived in Examples \ref{ex:genres} and \ref{ex:div}).

\hypertarget{weakening-strategies}{%
\section{Weakening Strategies}\label{weakening-strategies}}

As explained before, the weakening rule is mandatory in PB solvers to
maintain the inferred constraints conflictual. In the following, we
introduce different strategies for applying this rule in PB solvers,
designed towards finding a tradeoff between the strength of the inferred
constraints and their size.

\hypertarget{weakening-ineffective-literals-for-shorter-constraints}{%
\subsection{Weakening Ineffective Literals for Shorter
Constraints}\label{weakening-ineffective-literals-for-shorter-constraints}}

Within CDCL solvers, one captures the reason for a conflict being
encountered. A conflict occurs when a variable is propagated to both
\(0\) and \(1\). Intuitively, understanding why such a conflict occurred
amounts to understanding why these values have been propagated. In the
PB case, a constraint may be conflicting (resp. propagate literals) even
if it contains literals that are unassigned or already satisfied (see
Example \ref{ex:slack}). However, conflicts (resp. propagations) depend
only on \emph{falsified} literals (the slack of a constraint changes
only when one of its literals is falsified). Literals that are not
falsified are thus \emph{ineffective}: they do not play a role in the
conflict (resp. propagation), and may thus be weakened away. We can go
even further: when most literals are falsified, weakening some of them
may still preserve the conflict (resp. propagation), as shown in the
following example.

\begin{example}
\label{ex:noeffect}
Let $\rho$ be the partial assignment such that
$\rho(a) = \rho(c) = \rho(f) = 0$ (all other variables are unassigned).
Under $\rho$, $3\bar{a} + 3\bar{b} + c + d + e \geq 6$ has
slack $2$, so that literal $\bar{b}$ is propagated to $1$.
This propagation still holds after weakening away $\bar{a}$, $d$ and $e$,
giving after saturation $\bar{b} + c \geq 1$.
Similarly, consider the conflicting constraint $2a + b + c + f \geq 2$.
After the propagation of $\bar{b}$, weakening the constraint on $c$ and
applying saturation produces $a + b + f \geq 1$, which is still conflicting.
In both cases, the slack allows to detect whether a literal can be
weakened.
\end{example}

Observe that the constraints obtained are shorter, but are always
clauses. This guarantees that the resulting constraint will be
conflictual, but, if this operation is performed on both sides, only
clauses can be inferred, and the proof system boils down to resolution,
as in \emph{SATIRE} \cite{satire} or \emph{Sat4j-Resolution}
\cite{sat4j}.

\begin{example}[Example \ref{ex:noeffect} cont'd]
If a resolution step is performed between the weaker constraints
$\bar{b} + c \geq 1$ and $a + b + f \geq 1$, the clause $a + c + f \geq 1$ is
inferred.
However, if only one side is weakened, for example the conflict side, the
cancellation between $3\bar{a} + 3\bar{b} + c + d + e \geq 6$ and
$a + b + f \geq 1$ produces the constraint $3f + c + d + e \geq 3$.
Observe that, when the weakening operation is applied at the next step, the
stronger clause $c + f \geq 1$ is inferred after saturation.
\end{example}

\hypertarget{partial-weakening-for-stronger-constraints}{%
\subsection{Partial Weakening for Stronger
Constraints}\label{partial-weakening-for-stronger-constraints}}

To avoid the inference of constraints that are too weak to preserve the
strength of the proof system, an interesting option is to use
\emph{partial weakening}. Indeed, the weakening rule, as described
above, can be generalized as follows.

\begin{prooftree}
\AxiomC{$\alpha l + \sum_{i=1}^{n} \alpha_i l_i \geq \delta$}
\AxiomC{$\varepsilon \in \mathbb{N}$}
\AxiomC{$0 < \varepsilon \leq \alpha$}
\RightLabel{(partial weakening)}
\TrinaryInfC{$(\alpha - \varepsilon) l + \sum_{i=1}^{n} \alpha_i l_i \geq \delta - \varepsilon$}
\end{prooftree}

This rule is rarely used in practice by PB solvers, and the weakening
rule (i.e., the case when \(\varepsilon = \alpha\)) is often preferred.
However, partial weakening gives more freedom when it comes to inferring
new constraints, and allows to infer stronger constraints. We
implemented a variant of \emph{RoundingSat} which uses this rule as
follows. Before cancelling a literal out during conflict analysis, all
literals that are not currently falsified and have a coefficient not
divisible by the weight of the pivot are \emph{partially weakened}
(instead of simply \emph{weakened}). This operation is applied so that
the resulting coefficient becomes a multiple of the weight of the pivot.
This approach preserves the nice properties of \emph{RoundingSat} (see
\cite[Proposition 3.1 and Corollary 3.2]{rs}), and in particular the
fact that the produced constraint will be conflictual (the coefficient
of the pivot will be equal to \(1\)). It also allows to infer stronger
constraints, as illustrated by the following example.

\begin{example}
Let $\rho$ be the partial assignment defined by $\rho(a) = 1$ and
$\rho(b) = \rho(c) = \rho(d) = \rho(e) = 0$ (all other variables are
unassigned).
Consider the (conflicting) constraint $8a + 7b + 7c + 2d + 2e + f \geq 11$,
where $b$ is the literal to be cancelled out.
The above rule yields $7a + 7b + 7c + 2d + 2e \geq 9$ which, divided by $7$,
gives $a + b + c + d + e \geq 2$.
This constraint is stronger than the clause $b + c + d + e \geq 1$
inferred by \emph{RoundingSat}, which weakens away the literal $a$.
\end{example}

This variant has several advantages. First, its cost is comparable to
that of \emph{RoundingSat}: checking whether a coefficient is divisible
by the weight of the pivot is computed with the remainder of the
division of the former by the latter, which is the amount by which the
literal must be partially weakened. Second, the constraints it infers
may be stronger than that of \emph{RoundingSat}. Yet, this strategy does
not reduce the size of the constraints as much as the weakening of
ineffective literals. To get the best of both worlds, we introduce
tradeoff strategies.

\hypertarget{towards-a-tradeoff}{%
\subsection{Towards a Tradeoff}\label{towards-a-tradeoff}}

The previous sections showed that the weakening operation may help
finding short explanations for conflicts, but may also infer weaker
constraints. Several observations may guide us towards tradeoff
applications of the weakening rule.

First, the key property motivating \emph{RoundingSat} to round the
coefficient of the pivot to \(1\) does not require it to be equal to
\(1\) on \emph{both} sides of the cancellation: actually, having a
coefficient equal to \(1\) on only \emph{one} side is enough to
guarantee the resulting constraint to be conflicting \cite{dixon04}.
Weakening only the reason or the conflict is thus enough to preserve
this property, while maintaining coefficients low enough, as only one
side of the cancellation may need to be multiplied.

Second, one may apply the weakening rule in a different manner to keep
coefficients small so as to speed up arithmetic operations. A possible
approach is the following, that we call \emph{Multiply and Weaken}. Let
\(r\) be the coefficient of the pivot used in the cancellation appearing
in the reason and \(c\) that in the conflict. Find two values \(\mu\)
and \(\nu\) such that
\((\nu - 1) \cdot r < \mu \cdot c \leq \nu \cdot r\) (which can be done
using Euclidean division). Then, multiply the reason by \(\nu\), and
apply weakening operations on this constraint so as to
reduce the coefficient of the pivot to \(\mu\cdot c\). Note that, in
order to preserve the propagation, this coefficient cannot be weakened
directly. Instead, ineffective literals (as described above) are
successively weakened away so that the saturation rule produces the
expected reduction on the coefficient. Since this operation does not
guarantee to preserve the conflict, an additional weakening operation
has to be performed, as for generalized resolution. Note that this
approach may also derive clauses, even though this is not always the
case, as shown in the following example.

\begin{example}
Let $\rho$ be the partial assignment such that $\rho(a) = \rho(d) = 0$ and
$\rho(e) = 1$ (all other variables are unassigned).
Under $\rho$ the constraint $5a + 5b + 3c + 2d + e \geq 6$ propagates $b$.
The constraint $3\bar{b} + 2a + 2d + \bar{e} \geq 5$ becomes thus falsified.
Instead of using the LCM of $3$ and $5$ (i.e., $15$), the
reason of $b$ is weakened on $e$ and partially on $c$ to get, after saturation,
$3a + 3b + 2d + c \geq 3$.
The cancellation produces then $5a + 4d + c + \bar{e} \geq 5$.
\end{example}

\hypertarget{experimental-results}{%
\section{Experimental Results}\label{experimental-results}}

This section presents an empirical evaluation of the various strategies
introduced in this paper. To make sure that their comparison only takes
care of the underlying proof systems, and not of implementation details,
we integrated all of them in \emph{Sat4j} \cite{sat4j} (including
\emph{RoundingSat} proof system). The source code is available on
\href{https://gitlab.ow2.org/sat4j/sat4j/tree/weakening-investigations}{\emph{Sat4j}
repository}.

All experiments presented in this section have been run on a cluster
equiped with quadcore bi-processors Intel XEON E5-5637 v4 (3.5 GHz) and
128 GB of memory. The time limit was set to 1200 seconds and the memory
limit to 32~GB. The whole set of decision benchmarks containing
``small'' integers used in the PB competitions since the first edition
\cite{pb05} was considered as input.

\begin{figure}
\centering
\includegraphics[width=0.93\textwidth]{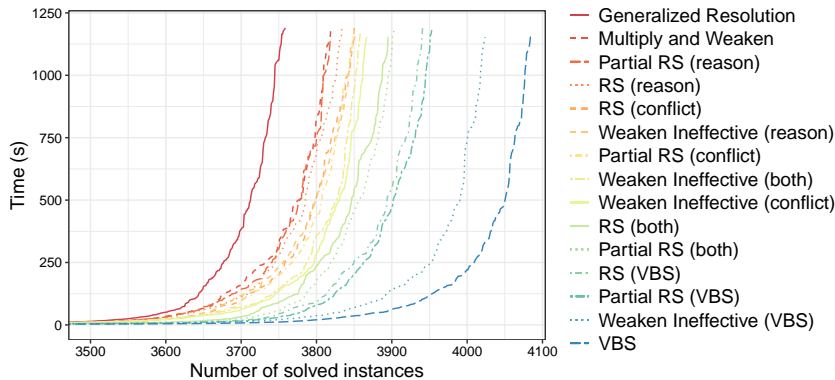}
\caption{Cactus plot of the different strategies implemented in Sat4j.
For more readability, the first 3,500 easy instances are cut out.
\label{fig:cactus}}
\end{figure}

As shown by Fig. \ref{fig:cactus}, strategies applying heavily the
weakening rule perform better than generalized resolution. Yet, among
these strategies, none of them is strictly better than the others. In
particular, the Virtual Best Solver (VBS), obtained by choosing the best
solver for each of the instances, performs clearly better than any
individual strategy. Each of these individual strategies does not have an important
contribution to the VBS, since the \emph{both}, \emph{conflict} and
\emph{reason} variants are very similar. However, if we consider the VBS
of the different ``main'' strategies, and in particular that of
\emph{RS}, \emph{Partial RS} and \emph{Weaken Ineffective} (also
represented on the cactus plot), their contributions become clearer:
\emph{Generalized Resolution} contributes to 6 instances, \emph{RS} to
13 instances, \emph{Partial RS} to 16 instances, and \emph{Weaken
Ineffective} to 83 instances. Even though \emph{Multiply and Weaken}
does not solve instances that are not solved by any other solver, it
solves 13 instances more than 1 second faster than any other approach (5
among them are faster solved by more than 1 minute). This suggests that
choosing the \emph{right} weakening strategy plays a key role in the
performance of the solver.

The strategies showing the best and most robust performance are those
applying the \emph{RoundingSat} (\emph{RS}) approach on both sides of
the cancellation rule, as they widely take advantage of the inference
power of the division rule. However, applying \emph{partial} weakening
instead of weakening gives better results, thanks to the stronger
constraints it infers. In particular, \emph{RS (both)} solves 3895
instances and \emph{Partial RS (both)} solves 3903 instances (with 3845
common instances). The performance of \emph{Partial RS (both)} is
evidenced on the \texttt{tsp} family, especially on satisfiable
instances: it solves 22 more such instances than \emph{RS (both)}, i.e.,
35 instances. For unsatisfiable instances, no common instances are
solved: \emph{Partial RS (both)} solves 7 instances, while \emph{RS
(both)} solves 5 distinct instances. In both cases, \emph{Partial RS
(both)} performs much more assignments per second than \emph{RS (both)},
allowing it to go further in the search space within the time limit.

Surprisingly, another strategy exhibiting good performance consists in
weakening ineffective literals on the \emph{conflict} side at each
cancellation (it contributes to 18 instances in the VBS). Similarly,
\emph{RoundingSat} strategies perform better when applied on the
conflict side rather than the reason side. Since the early development
of cutting planes based solvers, weakening has only been applied on the
\emph{reason} side (except for \emph{RoundingSat} \cite{rs}, which
applies it on \emph{both} sides). Our experiments show that it may be
preferable to apply it only on the conflict side: literals introduced
there when cancelling may still be weakened away later~on.

The gain we observe between the different strategies has several
plausible explanations. First, the solver does not explore the same
search space from one strategy to another, and it may thus learn
completely different constraints. In particular, they may be stronger or
weaker, and this impacts the size of the proof built by the solver.
Second, these constraints may be based on distinct literals, which may
have side effects on the VSIDS heuristic \cite{chaff}: different
literals will be ``bumped'' during conflict analysis. Such side effects
are hard to assess, due to the tight link between the heuristic and the
other components of the solver.

\hypertarget{conclusion}{%
\section{Conclusion}\label{conclusion}}

In this paper, we introduced various strategies for applying the
weakening rule in PB solvers. We showed that each of them may improve
the runtime of the solver, but not on all benchmarks. Contrary to the
approaches implemented in most PB solvers, the strategies consisting in
applying an aggressive weakening only on the conflict side provide
surprisingly good results. However, approaches based on
\emph{RoundingSat} perform better, but our experiments showed that
partial weakening is preferable in this context. This suggests that
weakening operations should be guided by the strength of the constraints
to infer. To do so, a perspective for further research consists in
searching for better tradeoffs in this direction.

\paragraph*{Acknowledgements.}
The authors are grateful to the anonymous reviewers for their numerous
comments, that greatly helped to improve the paper. Part of this work
was supported by the French Ministry for Higher Education and Research
and the Hauts-de-France Regional Council through the ``Contrat de Plan
État Région (CPER) DATA''.

\bibliographystyle{plain}
\bibliography{bibliography.bib}

\begin{thebibliography}{10}

\bibitem{galena}
Donald Chai and Andreas Kuehlmann.
\newblock {A fast pseudo-Boolean constraint solver}.
\newblock {\em {IEEE} Trans. on {CAD} of Integrated Circuits and Systems},
  pages 305--317, 2005.

\bibitem{cook87}
William Cook, Collette~R. Coullard, and György Tur\'{a}n.
\newblock {On the Complexity of Cutting-plane Proofs}.
\newblock {\em {Discrete Appl. Math.}}, pages 25--38, 1987.

\bibitem{dixon04}
Heidi Dixon.
\newblock {\em {Automating Pseudo-Boolean Inference Within a DPLL Framework}}.
\newblock PhD thesis, University of Oregon, 2004.

\bibitem{pbchaff}
Heidi~E. Dixon and Matthew~L. Ginsberg.
\newblock {Inference Methods for a Pseudo-Boolean Satisfiability Solver}.
\newblock In {\em {Proceedings of AAAI'02}}, pages 635--640, 2002.

\bibitem{minisat}
Niklas E{\'e}n and Niklas S{\"o}rensson.
\newblock {An Extensible SAT-solver}.
\newblock In {\em {Proceedings of SAT'04}}, pages 502--518, 2004.

\bibitem{minisatp}
Niklas Een and Niklas Sörensson.
\newblock {Translating Pseudo-Boolean Constraints into SAT}.
\newblock {\em {JSAT}}, pages 1--26, 2006.

\bibitem{rs}
Jan Elffers and Jakob Nordstr\"{o}m.
\newblock {Divide and Conquer: Towards Faster Pseudo-Boolean Solving}.
\newblock In {\em {Proceedings of IJCAI'18}}, pages 1291--1299, 2018.

\bibitem{gomory58}
Ralph~E. Gomory.
\newblock {Outline of an algorithm for integer solutions to linear programs}.
\newblock {\em {Bulletin of the American Mathematical Society}}, pages
  275--278, 1958.

\bibitem{haken85}
Armin Haken.
\newblock {The intractability of resolution}.
\newblock {\em {Theoretical Computer Science}}, pages 297--308, 1985.

\bibitem{hooker88}
John~N. Hooker.
\newblock {Generalized resolution and cutting planes}.
\newblock {\em {Annals of Operations Research}}, pages 217--239, 1988.

\bibitem{sat4j}
Daniel Le~Berre and Anne Parrain.
\newblock {The SAT4J library, Release 2.2, System Description}.
\newblock {\em {JSAT}}, pages 59--64, 2010.

\bibitem{pb05}
Vasco Manquinho and Olivier Roussel.
\newblock {The First Evaluation of Pseudo-Boolean Solvers (PB'05)}.
\newblock {\em {JSAT}}, pages 103--143, 2006.

\bibitem{grasp}
Joao Marques-Silva and Karem~A. Sakallah.
\newblock {GRASP: A Search Algorithm for Propositional Satisfiability}.
\newblock {\em {IEEE Trans. Computers}}, pages 220--227, 1999.

\bibitem{openwbo}
Ruben Martins, Vasco Manquinho, and In{\^e}s Lynce.
\newblock {Open-WBO: A Modular MaxSAT Solver}.
\newblock In {\em {Proceedings of SAT'14}}, pages 438--445, 2014.

\bibitem{chaff}
Matthew~W. Moskewicz, Conor~F. Madigan, Ying Zhao, Lintao Zhang, and Sharad
  Malik.
\newblock {Chaff: Engineering an Efficient SAT Solver}.
\newblock In {\em {Proceedings of DAC'01}}, pages 530--535, 2001.

\bibitem{nordstrom15}
Jakob Nordstr\"{o}m.
\newblock {On the Interplay Between Proof Complexity and SAT Solving}.
\newblock {\em {ACM SIGLOG News}}, pages 19--44, 2015.

\bibitem{handbook}
Olivier Roussel and Vasco~M. Manquinho.
\newblock {Pseudo-Boolean and Cardinality Constraints}.
\newblock In {\em {Handbook of Satisfiability}}, chapter~22, pages 695--–733.
  IOS Press, 2009.

\bibitem{naps}
Masahiko Sakai and Hidetomo Nabeshima.
\newblock {Construction of an ROBDD for a PB-Constraint in Band Form and
  Related Techniques for PB-Solvers}.
\newblock {\em {IEICE Transactions on Information and Systems}}, pages
  1121--1127, 2015.

\bibitem{pueblo}
Hossein~M. Sheini and Karem~A. Sakallah.
\newblock {Pueblo: A Hybrid Pseudo-Boolean SAT Solver}.
\newblock {\em {JSAT}}, pages 165--189, 2006.

\bibitem{jakobproofsolvers}
Marc Vinyals, Jan Elffers, Jes{\'{u}}s Gir{\'{a}}ldez{-}Cr\'{u}, Stephan Gocht,
  and Jakob Nordstr{\"{o}}m.
\newblock {In Between Resolution and Cutting Planes: {A} Study of Proof Systems
  for Pseudo-Boolean {SAT} Solving}.
\newblock In {\em {Proceedings of SAT'18}}, pages 292--310, 2018.

\bibitem{satire}
Jesse Whittemore and Joonyoung Kim Karem~A. Sakallah.
\newblock {{SATIRE:} {A} New Incremental Satisfiability Engine}.
\newblock In {\em {Proceedings of {DAC}'01}}, pages 542--545, 2001.

\end{thebibliography}

\end{document}